\pdfoutput=1

\documentclass[11pt]{article}

\usepackage[preprint]{acl}

\usepackage{times}
\usepackage{latexsym}

\usepackage[T1]{fontenc}

\usepackage[utf8]{inputenc}

\usepackage{microtype}

\usepackage{inconsolata}

\usepackage{graphicx}
\usepackage{multirow}
\usepackage{multicol}
\usepackage{booktabs}
\usepackage{tabularx}
\usepackage{amsmath}
\usepackage{amssymb}
\usepackage{color}
\usepackage{multicol}  
\usepackage{enumitem} 
\usepackage{float}  

\usepackage[most]{tcolorbox}
\usepackage{xcolor}
\newtcbox{\myshadowbox}[1][]{colframe=black!100!white, colback=yellow!10, boxrule=1pt, arc=4mm, auto outer arc, drop shadow, #1}
\title{Harnessing Large Language Models for Scientific Novelty Detection}

\author{
  Yan Liu\textsuperscript{1} \quad
  Zonglin Yang\textsuperscript{1} \quad
  Soujanya Poria\textsuperscript{2} \quad
  Thanh-Son Nguyen\textsuperscript{3} \quad
  Erik Cambria\textsuperscript{1}\thanks{Corresponding author: Erik Cambria.} \\
  \textsuperscript{1}Nanyang Technological University \\
  \textsuperscript{2}Singapore University of Technology and Design \\
  \textsuperscript{3}Agency for Science, Technology and Research (A*STAR) \\
  \texttt{\{yan010, zonglin.yang, cambria\}@ntu.edu.sg} \\
  \texttt{sporia@sutd.edu.sg, nguyen\_thanh\_son@ihpc.a-star.edu.sg} \\
}

\begin{document}
\maketitle
\begin{abstract}

In an era of exponential scientific growth, identifying novel research ideas is crucial and challenging in academia. Despite potential, the lack of an appropriate benchmark dataset hinders the research of novelty detection. More importantly, simply adopting existing NLP technologies, e.g., retrieving and then cross-checking, is not a one-size-fits-all solution due to the gap between textual similarity and idea conception. In this paper, we propose to harness large language models (LLMs) for scientific novelty detection (ND), associated with two new datasets in marketing and NLP domains. To construct the considerate datasets for ND, we propose to extract closure sets of papers based on their relationship, and then summarize their main ideas based on LLMs. To capture idea conception, we propose to train a lightweight retriever by distilling the idea-level knowledge from LLMs to align ideas with similar conception, enabling efficient and accurate idea retrieval for LLM novelty detection. Experiments show our method consistently outperforms others on the proposed benchmark datasets for idea retrieval and ND tasks. Codes and data are available at \url{https://anonymous.4open.science/r/NoveltyDetection-10FB/}.


\end{abstract}
\section{Introduction}

In the rapidly evolving landscape of scientific research, the ability to identify novel and underexplored ideas is critical for driving meaningful advancements. This challenge is particularly pronounced in specialized fields such as macadamia research, where the exponential growth of academic literature makes it difficult to discern truly original contributions \cite{zhao2025review}. In addition, despite the exponential growth in scientific and technological outputs, recent studies reveal a paradoxical decline in the novelty and disruptiveness of published papers and patents \cite{park2023papers}. Therefore, scientific novelty detection (ND) holds significant potential for accelerating innovation, but the absence of tailored benchmark datasets and underexplored methodologies limits progress in this area. 

To construct ND-tailored benchmark datasets, we propose to crawl a paper corpus with topological \textit{closure} in topological and \textit{compactness} for ND.  Specifically, the closure property indicates the completeness of the paper corpus for accurate ND, e.g., missing related papers make it possible to misclassify a non-novel idea into a novel one. To this end, we select a subset of papers as seed papers, based on which we extract their references into the corpus. Therefore, all the relevant papers included in the corpus become a closure set for these seed papers. To achieve the compactness for ND tasks, we elicit large language models (LLMs) to generate structured summaries of each paper’s core contributions, hypotheses, and methodologies, making it easy for these datasets to be utilized for ND tasks. We notice some datasets and benchmarks (e.g., DiscoveryBench~\cite{majumder2024discoverybench}) focus on content duplication, which is somewhat similar to ours. However, these datasets are either not closed-domain corpora or fail to concisely represent research ideas (e.g., the whole papers), limiting further exploration in this area.

\begin{figure}
    \centering
    \includegraphics[width=\linewidth]{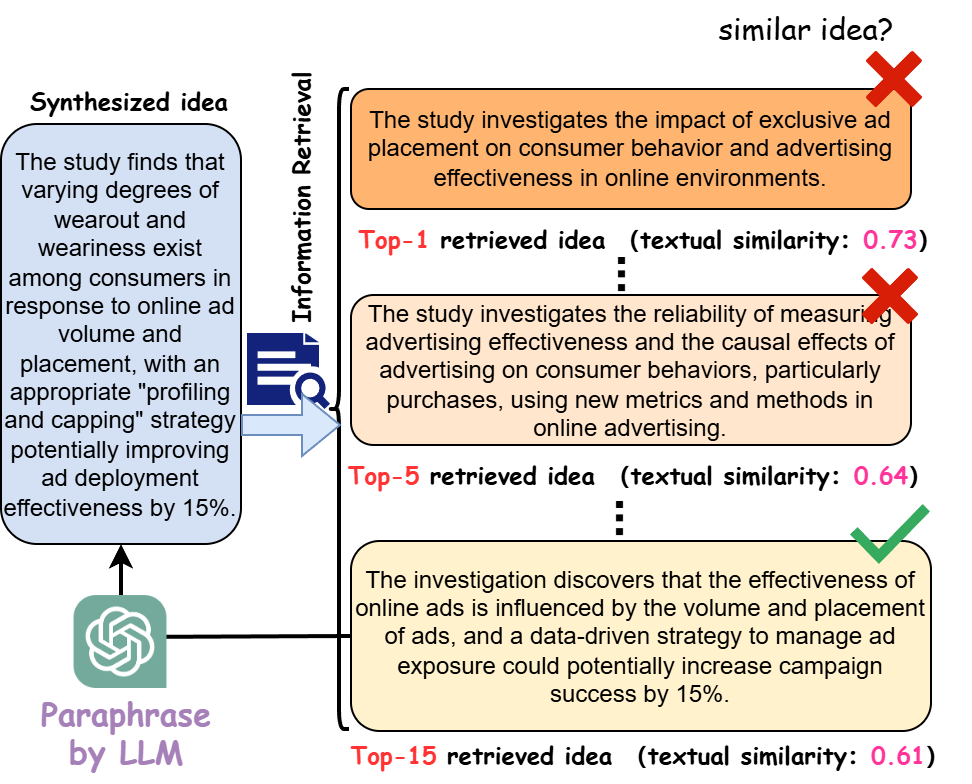}
    \caption{An illustrative example about the gap between textual similarity and idea conception. The LLM-paraphrased idea aligns conceptually with the target (green ticked) but shows lower textual similarity than distinct ideas retrieved by existing retrievers.}
    \label{fig:intro}
    \vspace{-20pt}
\end{figure}

Current ND methods predominantly rely on human expert assessments or heuristic measurements, which are resource-intensive and prone to biases arising from incomplete knowledge and the subjectivity of heuristic rule design ~\cite{jeon2023measuring,hofstra2020diversity,wang2017bias}. With the recent remarkable capabilities and rapid development of LLMs, it is intuitive to utilize their extensive knowledge, powerful text comprehension, and reasoning abilities \cite{zhao2023survey} to detect the novelty of research ideas. Intuitively, the LLM can easily identify its novelty if there is no similar idea in the corpus by cross-checking. Still, it is intractable to cross-check with all ideas in a large-scale corpus due to efficiency considerations. Therefore, the existing retrieval-augmented generation (RAG) strategy \cite{gao2023retrieval} is a promising way, i.e., to retrieve relevant ideas and then cross-check by LLMs, for both effective and efficient ND. However, simply utilizing RAG is not a one-size-fits-all solution for ND, leaving two main challenges for ND. Firstly, \textbf{\textit{the idea-level (conceptual) similarity is not well captured by existing retrievers}} compared to textual level similarity. \autoref{fig:intro} shows the textual level similarity of the target idea with respect to distinct ideas and synthesized ideas, where the synthesized idea is the paraphrase of the target idea by LLMs. Although the synthesized idea shares a similar idea conception to the target idea, it shares lower textual similarity compared to other distinct ideas identified by retrievers (e.g., BGE \cite{xiao2024c}). This greatly hinders the accuracy of ND by LLMs in the cross-checking phase. Secondly, \textbf{\textit{ there are no appropriate entities (e.g., non-novel paper) and relations (e.g., idea-to-idea pairs with similar academic ideas) to bridge between textual similarity and idea conception for retrievers}}. For example, the reference relation essentially implies task/tech-related relations rather than idea conception alignment. This leaves the idea-level alignment for the retriever unexplored.     

To address these challenges, we propose an LLM-based knowledge distillation (KD) framework to train an idea retriever, explicitly aligning ideas with similar conception by distilling knowledge from LLMs for ND. Firstly, to alleviate the absence of entities and relations for idea-level alignment, we generate a large-scale corpus of synthesized (non-novel) ideas based on anchor ideas by an LLM, which share overlapping conceptual content despite lower textual similarity. Specifically, to comprehensively model synthesized ideas by LLMs, we cover three types of generated ideas based on the information coverage of an anchor idea, namely rephrased ideas (information equivalence), partial ideas (information reduction), and incremental ideas (information addition). Secondly, we distill knowledge from LLMs into a lightweight retriever. Specifically, we propose fine-tuning a retriever based on idea pairs (anchor–synthesized ideas) to bridge textual similarity and idea conception. Finally, by cross-checking the target ideas with retrieved ideas using a tailored prompt, we can trigger LLMs to effectively and efficiently detect the novelty of a specific research idea. Our main contributions are threefold:
\begin{itemize}[leftmargin=*,itemsep=0pt, parsep=0pt]
 \item  We construct a largeND-tailored benchmark datasets for ND by ensuring the closure property and compactness.
 \item  {We propose an LLM-based ND framework that distills idea-level knowledge from LLMs into a lightweight retriever, using synthesized idea pairs to teach it to capture conceptual rather than textual similarity. This enables the framework to bridge the gap between surface-level text similarity and deeper idea-level alignment.}
 \item We conduct extensive experiments to validate the effectiveness of our framework in both the idea retrieval task and the ND task.
\end{itemize}

\section{Related work}
\paragraph{Novelty Detection} Existing novelty detection (ND) methods can be roughly categorized into three classes, namely citation–based methods, content-based methods, and multi-source-based methods. \textbf{citation network–based} methods aim to measure novelty by analyzing patterns and statistics in a paper’s citation graph \cite{uzzi2013atypical,lee2015creativity,wang2017bias,trapido2015novelty}. For example, \citet{uzzi2013atypical} introduced a Z-score metric that quantifies atypical journal-pair combinations in a paper’s reference list. \textbf{Content-based} methods propose to utilized different levels of granularity of textual data to quantify the novelty of scientific articles, much effort are devoted into exploring keywords \cite{azoulay2011incentives,mirowski2018future,yan2020impact,ruan2023effect}, entities \cite{liu2022pandemics,luo2022combination,chen2024exploring}, and sentences \cite{chen2019automatic,jeon2023measuring,wang2023measuring} via statistical or embedding-based metrics \cite{ruan2023effect,liu2022pandemics}. 
 \textbf{Multi-source-based} methods treat citations, entities, or concepts as evolving knowledge graphs and quantify novelty by structural perturbations \cite{shibayama2021measuring,amplayo2018network,hofstra2020diversity,de2021system}. For example, \citet{hofstra2020diversity} identified new cross-community links in concept co-occurrence networks as signals of disruptive or boundary-spanning work. However, these methods mainly rely on heuristic measurement strategies, which are resource-intensive and prone to biases arising from incomplete knowledge and the subjectivity of rule design. 

\paragraph{LLMs for Scientific Research} 
Due to the advantages of LLMs, recent studies have focused on generating scientific ideas and hypotheses with their assistance~\cite{li2024learning,wang2024scimon,yang2023large,lu2024ai,baek2024researchagent,liu2025researchbench}. Most of these methods adopt a generation-then-verification framework, aiming to produce novel ideas for scientific research. For example, SCIMON~\cite{wang2024scimon} encodes LLM-generated hypotheses using SentenceBERT, retrieves similar papers via vector search, computes cosine similarity scores, and refines the hypothesis until its similarity to existing work falls below a predefined novelty threshold. \citet{yang2023large} follows a comparable pipeline, integrating a “novelty checker” that iteratively optimizes hypotheses based on their cosine distance from prior literature. However, these novelty verification approaches rely solely on textual similarity measures, which fail to capture the core insights of an idea, leading to inaccurate assessments due to the gap between textual and conceptual similarity.

\section{Problem Formulation}

Suppose we have a novelty corpus $\mathcal{G}_N$ collected from real papers with $M$ novel ideas, and a synthesized corpus $\mathcal{G}_S$ generated by LLMs with $N$ non-novel ideas. Each idea $d \in \mathcal{G}_N \cup \mathcal{G}_S$ is associated with a textual description. For the novelty corpus $\mathcal{G}_N$, we assume that its ideas are novel and distinct from each other, i.e., {\small $\sum_{j \neq i,, d_j \in \mathcal{G}_N} I(d_i, d_j) = 0$} for each $d_i \in \mathcal{G}_N$, where $I(\cdot, \cdot)$ denotes the idea detector, such that $I(d_i, d_j) = 1$ if idea $d_i$ and idea $d_j$ are identical or similar, and 0 otherwise. For the synthesized corpus, we assume that its ideas are non-novel as they are generated based on $\mathcal{G}_N$, i.e.,  {\small $\sum_{d_j \in \mathcal{G}_N} I(d_i, d_j) \ge 1$} for each $d_i \in \mathcal{G}_S$.

In this work, we aim to research two kinds of tasks, namely the research \textit{\textbf{idea retrieval}} task and the \textbf{\textit{novelty detection}} (ND) task. In the idea retrieval task, given a synthesized idea $d\in \mathcal G_S$, our goal is to target the anchor papers from the novelty corpus, which can be formulated into retrieving a ranking list for the synthesized paper to hit the anchor papers. In novelty detection task, our goal is to check novelty of a given idea $d\sim \mathcal G_N\bigcup \mathcal{G}_S$, which can be novel (i.e., $d \in \mathcal G_N$) or non-novel (i.e., $d\in\mathcal G_S$). 

\section{Benchmark Dataset and Methodology}

\begin{figure*}
    \centering
    \includegraphics[width=\linewidth]{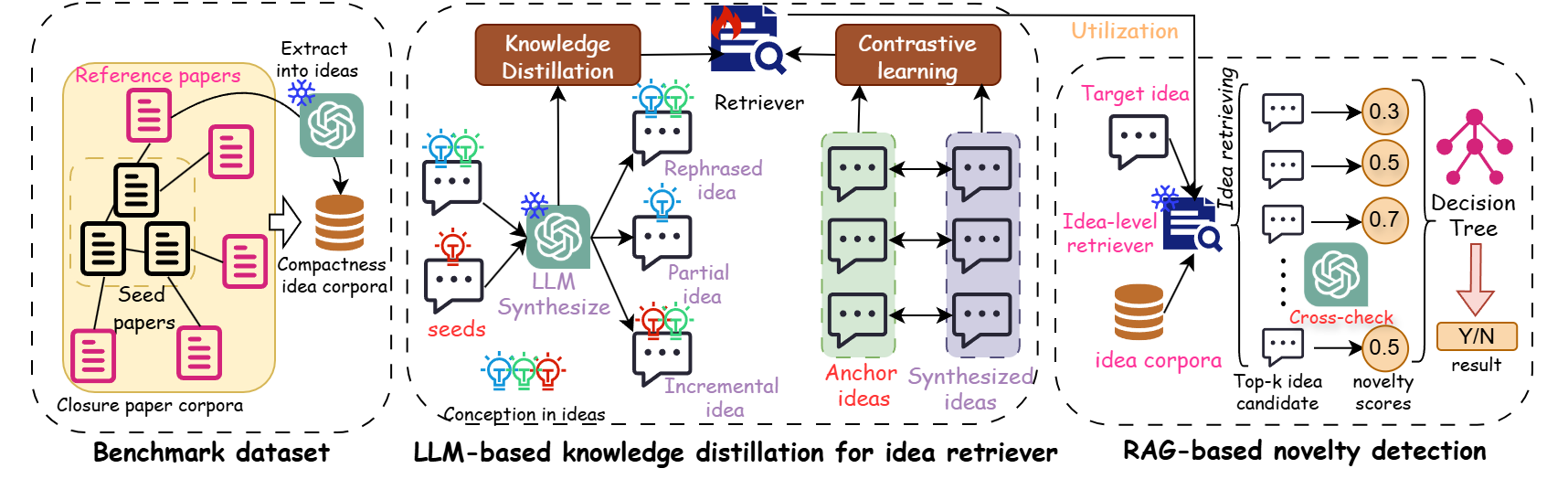}
    \caption{The overall architecture of the data construction and methodology. It includes three main components: (1) benchmark dataset construction using topological closure and idea compactness; (2) an LLM-based knowledge distillation framework to train the idea retriever using rephrased, partial, and incremental ideas; and (3) a RAG-based novelty detection strategy that retrieves top-k idea candidates and evaluates novelty scores for decision-making.}
    \label{fig:arch}
\end{figure*}

First, we introduce two ND-tailored benchmark datasets with topological closure and compactness for ND in Section \ref{sec:benchmark}. Second, we propose to train an idea retriever by an LLM-based KD framework in Section \ref{sec:KD}. Finally, we detail an ND strategy equipped with RAG in Section \ref{sec:ND}. The overall architecture of the data construction and methodology is shown in \autoref{fig:arch}.

\subsection{Benchmark datasets for ND}\label{sec:benchmark}
To construct ND-tailored benchmark datasets, we propose to crawl a paper corpus with consideration of the topological \textit{closure} and \textit{compactness} for ND.

To ensure the \textit{closure property} of the collected datasets, we propose selecting a subset of papers in a specific domain as \textit{seed papers}, based on which we extract their references into the corpus accordingly. Therefore, all the relevant papers included in the corpus form a closure set for these seed papers, i.e., no related paper published prior to the seed papers is excluded from the corpus. Specifically, we propose to extract seed papers from two representative research domains, 
\begin{itemize}[leftmargin=*,itemsep=0pt, parsep=0pt]
    \item \textbf{Marketing Domain.} Due to restrictions of social science publications, we collect 470 research articles from two leading journals in the Marketing field: the \textit{Journal of Marketing} and the \textit{Journal of Marketing Research}, spanning 2004 to 2024. 
    \item \textbf{Natural Language Processing (NLP) domain.} The NLP domain benefits from open-access practices. We systematically collected 3,533 papers from ACL conferences over the past five years to build the NLP dataset.
\end{itemize}
To collect references of seed papers into a closed corpus, we adopt the Semantic Scholar API \footnote{\url{https://www.semanticscholar.org/product/api}}, and collected reference papers for each seed paper, yielding a reference corpus of 12,832 papers in the Marketing domain and 33,911 in the NLP domain. 

\begin{tcolorbox}[breakable, colback=yellow!10, colframe=black!75!white, title={\small Prompt 2: Brief of synthesized idea generation}]\small
\textit{{\color{red} System}: You are an expert in scientific writing and language transformation. Your task is to paraphrase scientific statements while ensuring clarity, precision, and natural readability...
\newline {\color{red} User:}  {\color{magenta} Rephrased idea generation:} Given an input sentence, generate up to {k} sentences that retain the original information while allowing for modifications. ...
**Instructions:** ....
\newline {\color{red} User:}  {\color{magenta} Partial idea generation:} Given an input sentence, generate up to {k} subset sentences that retain part of the original information while allowing for minor modifications. You may add, delete, ... **Instructions:** ....
\newline {\color{red} User:}  {\color{magenta} Partial idea generation:} Given two input sentences, generate up to {k} sentences that retain the original information with no modifications or elaborations. The generated sentences are LAGRELY different from these two papers ... **Instructions:** ....
}
\end{tcolorbox}

\textbf{Idea extraction and summarization}
To achieve the {\textit{compactness property}} of corpora for ND tasks, we elicit LLMs to generate a summarization of each paper’s core contributions, hypotheses, and methodologies, making it easy for these datasets to be utilized for ND tasks. The detailed extract and summary rule can be found in Prompt 1 (a,b) in the Appendix. To ensure the effectiveness of idea extraction, we hire 3 experts (i.e., 2 Ph.D. students and 1 research fellow) to vote for alignment between of idea extraction and the original abstract on 50 papers by different LLM backbones (i.e., GPT-4o-mini, LLaMA3-3.1-8B, and PHI-3-3B). As a result, GPT-4o-mini showed optimal alignment and was chosen for idea extraction and summarization. 

\begin{figure*}[t] 
\centering
\begin{tcolorbox}[colback=yellow!10, colframe=black!89!white, 
  before skip=5pt, after skip=5pt,title={\small\textbf{Prompt 3:} Brief prompt of triggering LLMs for ND.}]\small  
\textit{{\color{red} System}: You are given a *"new research idea"* and a list of *"existing research ideas."* Your task is to compare the new idea **against each existing idea** and assign a **novelty score**.
\newline
{\color{red} User}: \#\#\# \textbf{Clarified Scoring Criteria:} Use the following novelty scoring rubric for each comparison. 
\newline \textbf{0.0 – No Novelty (Identical / Reworded)}. The new idea is a direct copy or a rephrased version of an existing idea.  Shares the same claims, findings, information or logic ...
\newline \textbf{0.3  – Low Novelty (Subset of Existing Idea)}. The new idea is a strict subset of an existing idea. Removes a condition, claim, or component, but otherwise shares the same logic and goal ...
\newline \textbf{0.5  – Moderate Novelty (large partial overlap)}. The new idea shares a substantial portion (around 50\%) of the ideas or claims with an existing idea. It introduces some new elements, such as ...
\newline \textbf{0.7  – High Novelty (small partial overlap)}. The new idea has minor similarities (e.g., method or framing), but differs in research focus, target, or core claim. It applies known ideas in a new context or formulates a distinct question ...
\newline \textbf{1.0 – Very High Novelty (Distinct Research Direction)}. The idea is entirely distinct in structure, claim, scope, and research objective. Only minor thematic or terminological similarity, if any ...
\newline \textbf{\#\#\# Evaluation Instructions:} For each comparison between the new idea and an existing idea: 1. Assess Overlap... 2. Assign a Novelty Score... 3. Repeat for All Comparisons...
\newline \#\#\# \textbf{Output Format:} [List of scores like: [0.3, 0.5, 0.3, 0.7, 1.0]].  Now, compare the given research idea with each of the existing ideas. For each comparison, assign a novelty score using the rubric above and no need to justify the score...
\newline {\color{magenta} Given Idea: {QUERY.}} {\color{blue} Existing Ideas: 1. \{Idea 1\}, ..., K. \{Idea K\}. }}
\end{tcolorbox}
\vspace{-10pt}
\end{figure*}

\subsection{LLM-based KD for idea retriever}\label{sec:KD}

Intuitively, LLMs can identify their novelty by depending on whether there is a similar idea in the corpus. However, the large-scale corpus makes it intractable to cross-check with all ideas, leaving it urgent to retrieve the most idea-level similar ideas. To train an idea retriever tailored for idea retrieval tasks, we propose an LLM-based KD framework that explicitly aligns ideas with similar conception through distilling knowledge from LLMs. To alleviate the absence of available entities and relations for idea alignment, we propose triggering LLMs to generate a large-scale corpus of synthesized ideas based on novelty ideas $\mathcal G_N$, then training a lightweight retriever for idea alignment. Specifically, we cover three types of synthesized ideas based on the information coverage of an anchor idea, i.e., equivalence, reduction, and addition.

\begin{itemize}[leftmargin=*,itemsep=0pt, parsep=0pt]
    \item \textbf{\textit{Rephrased idea}} is synthesized by rephrasing an idea using different linguistic expressions while preserving the original conceptions. This type of generation reflects \underline{information equivalence}, as it maintains the same research method, contribution, and hypothesis as the anchor idea.
    \item \textbf{\textit{Partial idea}} is synthesized by extracting a subset of the conception of an idea, such as isolating one specific contribution, methodology, or application domain.  This type of generation reflects \underline{information reduction}, where the synthesized idea represents an incomplete or narrowed version of the anchor idea.
    \item \textbf{\textit{Incremental idea}} is synthesized by extending the anchor idea with additional but closely related components, such as combining it with another idea or a minor extension. This represents \underline{information addition}, where the synthesized idea builds upon the original but includes additional conceptions in existing ideas.
\end{itemize}

We organize anchor-synthesized idea pairs into a base set $\mathcal{F} = \{(s_i, g_i) \mid s_i \in \mathcal{G}_N\}$, where $g_i = \text{LLM}(d_i, \texttt{Prompt2})$ is generated by LLMs based on different generation strategies (see brief in Prompt 2 and detail in Appendix Prompt 2 (a) (b) (c)).

To distill the knowledge from LLMs, we propose to fine-tune a retriever based on idea pairs \((s_i, g_i)\) to bridge between textual similarity and idea conception, where \(s_i \in \mathcal{G}_N\) is the anchor (novel) idea and \(g_i \in \mathcal{G}_S\) is the synthesized non-novel idea (i.e., rephrased, partial, or incremental). The core objective is to align the representation space of the retriever with the idea-level similarity determined by the LLM. Specifically, let \(f_{\theta}(\cdot)\) denote the embedding function of the retriever with initial parameters \(\theta\), which maps a text into a dense vector. We aim to train \(f_{\theta}\) such that the embedding of the synthesized idea \(g_i\) is close to its corresponding anchor \(s_i\), and far from unrelated novel ideas \(s_j \in \mathcal{G}_N, j \neq i\), which can be achieved via a contrastive learning objective:
\begin{equation*}
\mathcal{L} = \!\!\!\!\!\sum_{(s_i,g_i)\in \mathcal{F}} \!\!\!\!\!-\!\log \frac{\exp\left( \text{sim}(f_{\theta}(s_i), f_{\theta}(g_i)) / \tau \right)}{\sum_{j=1}^{|\mathcal{G}_N|} \exp\left( \text{sim}(f_{\theta}(s_j), f_{\theta}(g_i)) / \tau \right)}
\end{equation*}
where \(\text{sim}(\cdot, \cdot)\) is a similarity function (e.g., cosine similarity), and \(\tau\) is a temperature scaling factor.






In summary, the proposed KD framework ensures that the retriever learns to reflect idea-level similarity aligned with LLMs’ knowledge, rather than surface-level textual semantics. The lightweight retriever can be leveraged not only for assisting human-centric novelty evaluation, but also for efficiently retrieving relevant ideas to support downstream LLM-based ND.

\subsection{RAG-based novelty detection}\label{sec:ND}

After retrieving top-$K$ idea candidates from the corpus using the idea-level retriever, we perform a cross-checking procedure with LLMs to determine the novelty of the target research idea, addressing the core ND task by leveraging the reasoning capabilities and prior knowledge of LLMs. Specifically, given a target idea \( q \) and a set of retrieved candidate ideas \( \mathcal{C}_q=\{d_1, d_2, \ldots, d_K\} \subset \mathcal{G}_N\) by the idea retriever $f_{\theta}(\cdot)$, we design a structured prompt to guide the LLM to compare \( q \) against each \( d_i \) and output a novelty score compared to $d_i$,
\begin{equation*}
    s_q= \text{LLM}(q,\mathcal C_q,\texttt{prompt3})\in \mathbb{R}^K
\end{equation*}
where novelty scores are defined on 5 novelty levels, namely very high novelty, high novelty, moderate novelty, low novelty, and no novelty, which can be found in prompt 3 (see completed one in Appendix Prompt 3). 

Instead of relying on manually designed thresholds, we propose to learn the novelty decision rule directly from data via a supervised decision tree classifier. This approach captures non-linear combinations and interactions among novelty scores from retrieved ideas, enabling more flexible and accurate ND. Given a training dataset $\mathcal{D} = \{(s_q, y_q)\}$, where $y_q \in \{\texttt{Novel}, \texttt{Non-Novel}\}$ is the ground truth label ($y_q=\texttt{Novel}$ if $q\in\mathcal G_N$ and $y_q=\texttt{Non-Novel}$ if $q\in\mathcal G_S$), we train a decision tree classifier $DTree(\cdot)$, therefore the final decision about ND can be formulated by the decision tree classifier $\hat{y}_q = DTree(s_q)$.

\section{Experiments}

In this section, we aim to validate the effectiveness of the proposed method. Specifically, we conduct extensive experiments in both idea retrieval and the ND task to study the following research questions:  \textbf{RQ1}: Whether the existing retrievers benefit from the proposed LLM-based KD framework in idea retrieval tasks? \textbf{RQ2}: Whether the proposed method benefits from the idea retriever in ND tasks? 
\textbf{RQ3}: How do hyperparameters influence the performance of the proposed method?

\subsection{Experimental Setup}
 \paragraph{Datasets.} We utilized the newly proposed benchmark datasets in both Marketing and NLP domains, described in Section 3.1. Specifically, the \textbf{\textit{Marketing}} dataset comprises 470 seed papers and their closure references totaling 12,577 unique papers. The \textbf{\textit{NLP}} dataset contains 3,533 seed papers and their associated closure references, totaling 32,239 unique papers. Negative examples (non-novel ideas) were generated from seed papers using GPT-4o-mini with rephrased, partial, and incremental prompts, producing up to 10 synthesized variants per anchor paper.  To prevent data leakage, we propose filtering each reference corpus based on the publication date of its corresponding seed paper. An overlap of 255 (Marketing) and 1,672 (NLP) papers between seed and reference sets was identified and removed appropriately. 
\subsection{Experiments on idea retrieval tasks (RQ1)}

\paragraph{Baseline Methods.} We compared our proposed distilled retriever with several state-of-the-art retrievers:
 \textbf{\textit{Vanilla}}: The vanilla version of the retriever backbone without fine-tuning. \textit{\textbf{Reference alignment (RA)}}: Fine-tuning the retriever backbone based on anchor-reference alignment. \textit{\textbf{LLM-based knowledge distillation retriever (LLM KD)}}: Fine-tuning the retriever backbone based on synthesized ideas generated by LLMs.

\paragraph{Retriever Backbone.} To comprehensively evaluate the proposed LLM-based KD framework, we adopt 6 well-known retriever backbones for comparison, namely General Text Encoder (GTE) \cite{li2023towards}, E5 \cite{wang2022text}, SimCSE \cite{gao2021simcse}, Sentence-BERT Paraphrase (SBERT\_p) \cite{reimers2019sentence}, Natural Language Inference-tuned (NLI) \cite{conneau2017supervised}, and Baidu General Embedding (BGE) \cite{xiao2024c}.

\paragraph{Evaluation Protocol, Metrics, and Implementation Details.} We split the seed papers and their corresponding negative examples into train, valid, and test subsets by a ratio of 6:1:3.  To evaluate idea retrieval performances of different models, we adopt standard retrieval metrics in IR tasks: Acc@k and MAP, where we choose $k\in\{1,5,10,20,50\}$ empirically. For the retriever fine-tuning, we set the learning rate of $2e^{-5}$, batch size of 16, and a cosine similarity function with temperature scaling. 


\paragraph{Performance Comparison.} \autoref{table:main1} presents the comparative results in idea retrieval task. From the experimental results, we obtain the following conclusions: Firstly, LLM-based KD retriever consistently outperforms baseline methods across both domains, which shows the effectiveness of the proposed method. LLM-based KD retriever achieves notable enhancement, with average improvements of 5.40\% and 15.19\% when compared to the top-performing baseline method on the Marketing domain and NLP task. Secondly, the variant RA, which is trained by anchor-reference alignment, degrades the Vanilla retriever in most cases, showing that anchor-reference usually excludes idea-level similarity. Specifically, although the anchor papers and their reference papers usually share a similar research question and background, the ideas and novel conception of them are distinctive as the novelty requirement for publication. Thirdly, the proposed method LLM-based KD retriever achieves more improvement on the NLP dataset with more papers compared to Vanilla. This is attributed to the large-scale papers in the corpus making the synthesized idea be less likely to be retrieved based on textual similarity, but it can be effectively retrieved based on idea similarity.

\renewcommand{\arraystretch}{1.6}
\begin{table*}[t]
  \centering
  \fontsize{8}{6}\selectfont
  \caption{{\small Experiments idea retrieval with different embedding backbones. The proposed LLM-KD retriever outperforms RA (Reference Alignment) and Vanilla (baseline without supervision) baselines, demonstrating the effectiveness of idea-level supervision.}}\label{table:main1}
  \vspace{-10pt}
\begin{tabular}{|c|c|ccccc|ccccc|}
\midrule\multicolumn{2}{|c|}{Domain}               & \multicolumn{5}{|c|}{Marketing}                                                           & \multicolumn{5}{c|}{NLP}                                                                      \\ \midrule
\multicolumn{1}{|l|}{Backbone} & Methods      & Acc@5           & Acc@10          & Acc@20          & Acc@50          & MAP             & Acc@5           & Acc@10          & Acc@20          & Acc@50          & MAP             \\\midrule
\multirow{4}{*}{GTE}         & Vanilla      & 0.7525          & 0.7907          & 0.8303          & 0.8612          & 0.6541          & 0.5585          & 0.5804          & 0.6003          & 0.6248          & 0.5025          \\
& RA           & 0.6847          & 0.7257          & 0.7589          & 0.8089          & 0.5963          & 0.4747          & 0.5001          & 0.5229          & 0.5500          & 0.4289          \\
& LLM-KD & \textbf{0.7593} & \textbf{0.8021} & \textbf{0.8321} & \textbf{0.8662} & \textbf{0.6649} & \textbf{0.5846} & \textbf{0.6096} & \textbf{0.6294} & \textbf{0.6578} & \textbf{0.5265} \\
& Imprv.       & 0.90\%          & 1.44\%          & 0.22\%          & 0.58\%          & 1.65\%          & 4.67\%          & 5.02\%          & 4.84\%          & 5.29\%          & 4.79\%          \\\midrule
\multirow{4}{*}{E5}          & Vanilla      & 0.7357          & 0.7712          & 0.8035          & 0.8412          & 0.6370          & 0.5552          & 0.5766          & 0.5974          & 0.6238          & 0.5014          \\
& RA           & 0.6879          & 0.7261          & 0.7589          & 0.8203          & 0.6037          & 0.4875          & 0.5142          & 0.5378          & 0.5674          & 0.4374          \\
& LLM-KD & \textbf{0.7530} & \textbf{0.7925} & \textbf{0.8217} & \textbf{0.8640} & \textbf{0.6566} & \textbf{0.5828} & \textbf{0.6085} & \textbf{0.6306} & \textbf{0.6561} & \textbf{0.5249} \\
& Imprv.       & 2.35\%          & 2.76\%          & 2.27\%          & 2.71\%          & 3.08\%          & 4.98\%          & 5.53\%          & 5.55\%          & 5.18\%          & 4.70\%          \\\midrule
\multirow{4}{*}{SimCSE}      & Vanilla      & 0.5905          & 0.6356          & 0.6783          & 0.7416          & 0.5061          & 0.3849          & 0.4113          & 0.4364          & 0.4662          & 0.3482          \\
& RA           & 0.5896          & 0.6442          & 0.6906          & 0.7530          & 0.5323          & 0.3897          & 0.4215          & 0.4451          & 0.4817          & 0.3475          \\
& LLM-KD & \textbf{0.6479} & \textbf{0.6888} & \textbf{0.7320} & \textbf{0.7916} & \textbf{0.5575} & \textbf{0.5459} & \textbf{0.5730} & \textbf{0.5981} & \textbf{0.6322} & \textbf{0.4869} \\
& Imprv.       & 9.72\%          & 6.92\%          & 5.99\%          & 5.13\%          & 4.73\%          & 40.09\%         & 35.93\%         & 34.37\%         & 31.24\%         & 39.84\%         \\\midrule
\multirow{4}{*}{sbert\_p}    & Vanilla      & 0.6037          & 0.6465          & 0.6911          & 0.7571          & 0.5338          & 0.4919          & 0.5197          & 0.5457          & 0.5769          & 0.4414          \\
& RA           & 0.6447          & 0.6947          & 0.7407          & 0.7966          & 0.5727          & 0.4509          & 0.4823          & 0.5088          & 0.5440          & 0.4002          \\
& LLM-KD & \textbf{0.6793} & \textbf{0.7243} & \textbf{0.7775} & \textbf{0.8153} & \textbf{0.5825} & \textbf{0.5500} & \textbf{0.5761} & \textbf{0.6009} & \textbf{0.6344} & \textbf{0.4902} \\
& Imprv.       & 5.37\%          & 4.26\%          & 4.97\%          & 2.34\%          & 1.71\%          & 11.80\%         & 10.85\%         & 10.12\%         & 9.96\%          & 11.06\%         \\\midrule
\multirow{4}{*}{NLI}         & Vanilla      & 0.3558          & 0.3935          & 0.4327          & 0.4959          & 0.3066          & 0.2085          & 0.2259          & 0.2440          & 0.2678          & 0.1871          \\
& RA           & 0.6160          & 0.6592          & 0.6938          & 0.7439          & 0.5534          & 0.4411          & 0.4695          & 0.4951          & 0.5316          & 0.3994          \\
& LLM-KD & \textbf{0.7056} & \textbf{0.7548} & \textbf{0.7934} & \textbf{0.8430} & \textbf{0.6151} & \textbf{0.5613} & \textbf{0.5874} & \textbf{0.6084} & \textbf{0.6425} & \textbf{0.5042} \\
& Imprv.       & 14.54\%         & 14.50\%         & 14.35\%         & 13.33\%         & 11.14\%         & 27.23\%         & 25.10\%         & 22.89\%         & 20.87\%         & 26.23\%         \\\midrule
\multirow{4}{*}{BGE}         & Vanilla      & 0.7266          & 0.7698          & 0.7994          & 0.8312          & 0.6339          & 0.5294          & 0.5543          & 0.5749          & 0.6023          & 0.4735          \\
& RA           & 0.7093          & 0.7425          & 0.7743          & 0.8248          & 0.6154          & 0.4816          & 0.5074          & 0.5291          & 0.5581          & 0.4319          \\
& LLM-KD & \textbf{0.7675} & \textbf{0.8089} & \textbf{0.8380} & \textbf{0.8703} & \textbf{0.6636} & \textbf{0.5812} & \textbf{0.6047} & \textbf{0.6269} & \textbf{0.6587} & \textbf{0.5225} \\
& Imprv.       & 5.63\%          & 5.08\%          & 4.83\%          & 4.70\%          & 4.69\%          & 9.79\%          & 9.10\%          & 9.04\%          & 9.37\%          & 10.35\%     \\\midrule   
\end{tabular}
\vspace{-10pt}
\end{table*}

\renewcommand{\arraystretch}{1.6}
\begin{table}[t]
  \centering    
  \caption{{\small Experiments on different types of synthesized (non-novel) ideas with BGE backbone in NLP dataset. }}\label{table:group}
  \fontsize{8}{6}\selectfont
\begin{tabular}{|c|c|c|c|c|}\midrule   
Type  & Metric & Vanilla & LLM KD & Imprv.  \\\midrule  
\multirow{3}{*}{\begin{tabular}[c]{@{}c@{}}Rephrased\\  idea\end{tabular}}   & Acc@5  & 0.9815  & 0.9887 & 0.74\%  \\
& Acc@20 & 0.9899  & 0.9952 & 0.53\%  \\
& MAP    & 0.9667  & 0.9767 & 1.04\%  \\\midrule
\multirow{3}{*}{\begin{tabular}[c]{@{}c@{}}Partial\\  idea\end{tabular}}     & Acc@5  & 0.9296  & 0.9554 & 2.77\%  \\
& Acc@20 & 0.9618  & 0.9787 & 1.76\%  \\
& MAP    & 0.8911  & 0.9230 & 3.58\%  \\\midrule
\multirow{3}{*}{\begin{tabular}[c]{@{}c@{}}Incremental \\ idea\end{tabular}} & Acc@5  & 0.3404  & 0.4079 & 19.82\% \\
& Acc@20 & 0.3970  & 0.4672 & 17.67\% \\
& MAP    & 0.2715  & 0.3329 & 22.61\% \\\midrule
\end{tabular}
\vspace{-10pt}
\end{table}

\renewcommand{\arraystretch}{1.6}
\begin{table*}[t]
  \centering
  \fontsize{8}{6}\selectfont
  \caption{{\small Experiments on different methods in ND task. The proposed RAG-KD method outperforms all baselines, as demonstrated by the percentage gains. }}\label{table:main2}
  \vspace{-10pt}
\begin{tabular}{|c|cccc|cccc|}\midrule
Dataset     & \multicolumn{4}{|c|}{Marketing}                                         & \multicolumn{4}{|c|}{NLP}                                               \\\midrule
Baseline    & Accuacy         & Precision       & Recall          & F1              & Accuacy         & Precision       & Recall          & F1              \\\midrule
CD          & 0.6005          & 0.6395          & 0.6005          & 0.5797          & 0.5167          & 0.5406          & 0.5167          & 0.4430          \\
URPC        & 0.5500          & 0.5565          & 0.5500          & 0.5366          & 0.4800          & 0.3698          & 0.4800          & 0.3404          \\
PES         & 0.4700          & 0.4577          & 0.4700          & 0.4283          & 0.5800          & 0.5891          & 0.5800          & 0.5690          \\
MOOSE       & 0.6400          & 0.7604          & 0.6400          & 0.5929          & 0.5900          & 0.6018          & 0.5900          & 0.5778          \\
SciMON      & 0.4900          & 0.4745         & 0.4900          & 0.3985         & 0.6100          & 0.6244          & 0.6100          & 0.5984          \\\midrule
RAG-Vanilla & 0.7292          & 0.7664          & 0.7292          & 0.7180          &  0.6100  &0.63356&    0.61000 &   0.59201       \\
RAG-KD      & \textbf{0.7453} & \textbf{0.8023} & \textbf{0.7453} & \textbf{0.7344} & \textbf{0.7474} & \textbf{0.8010 } & \textbf{ 0.7474   } & \textbf{0.7349} \\\midrule
Improve     & 24.11\%         & 25.46\%         & 24.11\%         & 26.69\%         & 22.54\%	&28.29\% &	22.54\%	&22.82\%       \\\midrule
\end{tabular}
\vspace{-10pt}
\end{table*}

\paragraph{Group Analysis on Synthesized Ideas.}
To further investigate the effectiveness of the proposed LLM-based knowledge distillation, we analyze its performance across different types of synthesized (non-novel) ideas, including \textit{Rephrased}, \textit{Partial}, and \textit{Incremental} ideas. As shown in Table~\ref{table:group}, we observe consistent improvements across all types when using the LLM-based KD retriever compared to the vanilla BGE retriever. Most significantly, the KD retriever demonstrates the largest improvement on incremental ideas, which have the lowest textual similarity but share similar ideas with their anchors. This highlights that the LLM-based KD retriever can effectively capture the idea-level similarity. For example, the ``incremental contribution and novelty'' is usually recognized as the main weakness in reviews by experts for submission. 

\subsection{Experiments on idea retrieval tasks (RQ2)}

\paragraph{Baseline Methods} For ND, we compared the proposed methods with several state-of-the-art baselines and some variants of our method: \textbf{URPC}: \citet{uzzi2013atypical} propose measure novelty by identifying \underline{u}nusually \underline{r}are journal‐\underline{p}air \underline{c}ombinations in a paper’s reference list. \textbf{PES}: \citet{liu2022pandemics} detect novelty in COVID-19 research by measuring the \underline{p}roportion of biological \underline{e}ntity pairs with high \underline{s}emantic distance using BioBERT embeddings. \textbf{CD}: \citet{shibayama2021measuring} quantify novelty by computing \underline{c}osine \underline{d}istances among word embeddings of a paper’s cited references. \textbf{SCIMON}: \citet{wang2024scimon} propose an idea generation framework for scientific research. It heuristically judges the novelty based on a threshold of cosine similarity to existing ideas. \textbf{MOOSE}: \cite{yang2024moose} propose an idea generation framework for scientific research. It directly triggers LLMs to judge the novelty of a specific idea. \textbf{RAG-Vanilla/KD}: Our proposed \underline{RAG}-based ND by LLMs, where the we adopt the \underline{Vanilla}/LLM-based \underline{KD} retriever for RAG. 

\paragraph{Evaluation Protocol, Metrics, and Implementation Details.}
For efficient evaluation, we randomly select 100 samples from the train (for decision tree training) and test sets (for evaluation) with 1:1 novel and non-novel instances, respectively. As ND is a classification tasks, we adopt accuracy, precision, recall, and F1-score to evaluate binary classification performance. We adopt the LLM-based KD BGE as the RAG retriever due to its effectiveness for idea retrieval, employing deepseek-reasoner as the LLM backbone for the ND task. We retrieve top-$10$ and top-$5$ ideas as candidate ideas of RAG in Marketing and NLP datasets, respectively.

\begin{figure}
    \centering
    \includegraphics[width=0.85\linewidth]{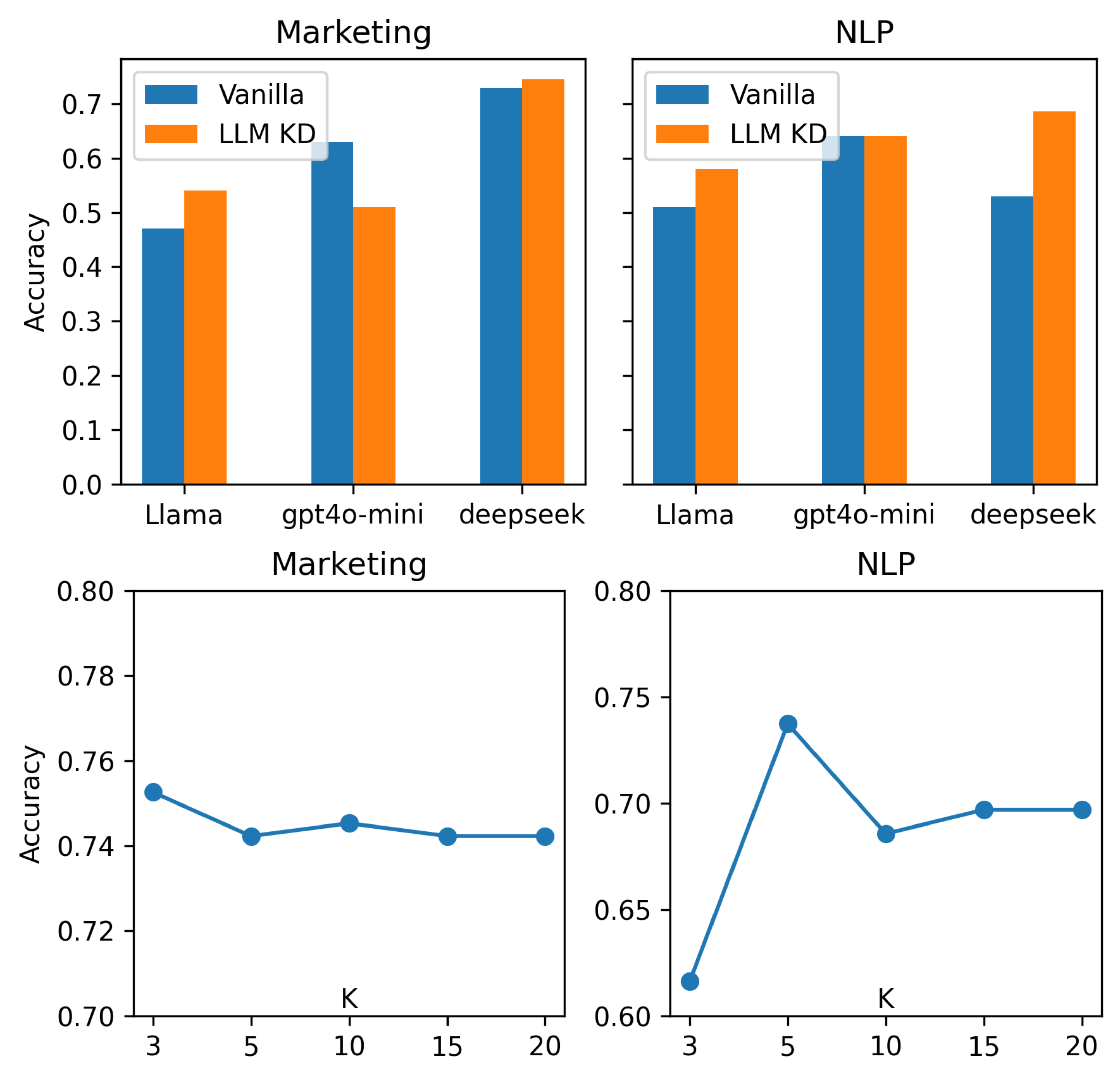}
    \caption{Investigate on (a) LLM backbones and (b) retrieval size $K$ for ND tasks .}
    \label{fig:hyper}
\vspace{-10pt}
\end{figure}

\paragraph{Performance comparison} \autoref{table:main2} presents the comparative results in ND task. From the experimental results, we obtain the following conclusions: Firstly, our method RAG-KD consistently outperforms baseline methods across both domains, which shows the effectiveness of the proposed method. Secondly, RAG-KD outperforms RAG-Vanilla in all cases, showing the importance of accurately retrieving conceptional similar ideas for ND tasks. Thirdly, among baseline methods, MOOSE achieves a competitive performance among these methods, indicating the effectiveness of utilizing LLMs for novelty detection. Several baselines (e.g., SciMON, PES, CD) show huge performances varying with domains, indicating that heuristic rules may be optimal for all scenarios.

\subsection{Hyperparameter analysis (RQ3)}
In this subsection, we further conduct experiments to analyze the impact of hyperparameters and the LLM backbone selection for ND tasks.
\paragraph{LLM Backbone.}
 \autoref{fig:hyper} (a) evaluates our method with different backbones, namely Llama-3.1-8B-Instruct, gpt4o-mini, and deepseek-reasoner. Firstly, the LLM-based KD retriever outperforms the Vanilla retriever in most cases, validating the general applicability of our framework. Secondly, the deepseek-reasoner consistently outperforms other LLM backbones. We suggest employing the deepseek-reasoner with an LLM-based KD retriever for real-world implementation for ND tasks.

\paragraph{Retrieval Size $K$.} \autoref{fig:hyper} (b) investigates the influence of the number of retrieved ideas \( K \), indicating that moderate $K$ (e.g., 5 and 10) contributes to stable and optimal performances of our method. We notice that the large $K$ (e.g., 20) does not promise the optimal performances, which may be attributed to the limited capability of LLM for handling large-scale ideas~\cite{liu-etal-2024-lost}.

\section{Conclusion}

In this paper, we propose two ND-tailored benchmark datasets, namely Marketing and NLP, for ND by ensuring the closure property and compactness. To extract the knowledge and reasoning capability of LLMs into a retriever, we propose an LLM-based knowledge distillation framework to elicit the retriever to identify potential idea-similar in the corpus, enabling accurate idea retrieval and ND tasks. Finally, we propose a cross-checking procedure with LLMs to determine the novelty of the target research idea.
Extensive experiments validate the effectiveness of our framework in both the idea retrieval task and the ND task.

\section{Limitation}
The primary constraints of this paper are as follows: 
1) The LLM-generated ideas and novelty scores are not guaranteed to be fully accurate or consistent, especially when the source prompts are subtle or ambiguous. Such noise in pseudo labels may affect the quality of retriever fine-tuning and ND.
2) Our framework currently models ND as a binary classification task. However, novelty is often subjective and continuous, which may require future extensions to soft or human-in-the-loop evaluations.

\bibliography{main}
\newpage
\appendix
\section{Appendix}\label{sec:appendix}
\begin{figure}[t]
\vspace{-20pt}
\vspace{-1em}
  \begin{tcolorbox}[colback=yellow!10, colframe=black!89!white, before skip=5pt, after skip=5pt,
      title=\textbf{Prompt 1 (b):} Detailed prompt for idea extraction in Marketing domain.,
      fontupper=\small,               
      sharp corners,
      ]  
\textit{{\color{red} System}: You are a social science expert searching for new research topics to explore. To accomplish this, you need to conduct a comprehensive literature review to understand the current research landscape, including the challenges being addressed, the hypotheses being tested, the methodologies employed, and the specific conditions under which key findings are obtained.
\newline
{\color{red} User}: In social science research, a hypothesis is typically a clear, specific, and testable statement predicting the relationship between an independent variable and a dependent variable. It often outlines how one variable influences another and to what degree. The hypothesis is central to the research, as it guides the study's aim of proving or disproving this relationship. In many cases, the hypothesis reflects the main conclusion or finding of the research. Keywords or key phrases are essential terms capturing the study's core concepts, variables, and major themes, and should reflect the primary focus areas. They should neither be too specific nor too general, avoiding keyword over-expansion while effectively indexing the research.
\newline Example of Research Hypotheses in Social Science: Viewing a visually depicted product that facilitates embodied mental simulation leads to heightened purchase intentions, with perceptual resources for mental simulation attenuating this effect, and for negatively valenced products, it decreases purchase intentions.
\newline Your task is to extract the hypothesis from each title and abstract pair of a research paper. 
\newline Given time constraints, you only need to review the title and abstract of research papers published in top-tier journals. Your task is to extract the hypothesis from each title and abstract pair. The hypothesis should be comprehensive, including key details or side hypotheses if present. If no hypothesis is found, you can extract the findings or conclusions of the contributions of the study as the hypothesis.
\newline Below is the target article:
\newline Title: {\color{magenta} \{title\}}
\newline Abstract: {\color{magenta} \{abstract\}}
\newline After reviewing the abstract, please extract the main hypothesis using the following format (only the hypothesis, no additional formatting, no additional text to explain your answer):
\newline  Hypothesis: [Extracted hypothesis, including details if present, put value to be 'None' if no hypothesis is found]\#
}
\end{tcolorbox}
\end{figure}

\vspace{43pt}
  \begin{tcolorbox}[colback=yellow!10, colframe=black!89!white, before skip=5pt, after skip=5pt,
      title=\textbf{Prompt 1 (a):} Detailed prompt for idea extraction in NLP domain.,
      fontupper=\small,               
      sharp corners,
      ]  
{{\color{red} System}: You are a computer science expert searching for new research ideas. To accomplish this, you need to conduct a comprehensive literature review to understand the current research landscape, including the research tasks, the methods used on the tasks, and the specific conditions under which key findings are obtained.
\newline
{\color{red} User}: A research hypothesis in a computer science paper is a testable statement or prediction about a specific phenomenon, system, or algorithm that the research aims to investigate or validate. It serves as the foundation of the study, guiding the research design, experiments, and analysis. The hypothesis should be clearly articulated and typically stems from prior knowledge, gaps in existing literature, or a novel idea the authors aim to explore. A good hypothesis should be specific, testable, context-dependent, and relevant to the research problem. When extracting the hypothesis, ensure it begins directly with the subject of the statement (e.g., "Optimistic posterior sampling algorithm for reinforcement learning (OPSRL)...") without introductory phrases such as "The proposed" or "We propose." Focus on presenting the hypothesis succinctly and directly. Examples of Research Hypotheses in Computer Science: Algorithm Development: "METHODS will perform better in terms of computational efficiency and accuracy compared to existing algorithms for large-scale data sorting." Machine Learning: "Incorpusting domain-specific embeddings in the neural network architecture will significantly improve its performance on task X."
\newline Your task is to extract the hypothesis from each title and abstract pair of a research paper.  If you cannot find any hypothesis in the abstract, you can extract the findings or conclusions of the contributions of the study as the hypothesis. If the input abstract is not meaningful, you can output 'None'.
\newline Below is the target article:
\newline Title: {\color{magenta} \{title\}}
\newline Abstract: {\color{magenta} \{abstract\}}
\newline After reviewing the abstract, please extract the hypothesis using the following format (only the hypothesis, no additional formatting, no additional text to explain your answer):
\newline  Hypothesis: [Extracted hypothesis, including details if present, put value to be 'None' if no hypothesis is found]\#
}
\end{tcolorbox}

\begin{figure*}[t]
  \centering
  \begin{tcolorbox}[colback=yellow!10, colframe=black!89!white, before skip=5pt, after skip=5pt,
      title=\textbf{Prompt 2 (a):} Detailed prompt for rephrased idea generation by LLMs,
      fontupper=\small,               
      sharp corners,
      ]  
{{\color{red} System}: You are an expert in scientific writing and language transformation. Your task is to paraphrase scientific statements while ensuring clarity, precision, and natural readability. The meaning of the original sentence must remain unchanged, but the structure and expression should be significantly altered.
\newline
{\color{red} User}: 
\newline \#\#\#\#**Task:** Given an input sentence, generate **up to 5 sentences** that retain the original information while allowing for modifications. Each generated sentence should be of the same meaning as the original, but with completely different words and structure.
\newline  \#\#\#\# **Instructions:**  
\newline 1. **Change Words to Synonyms**  
\newline - Replace key terms with scientifically appropriate synonyms while maintaining clarity and rigor.  
\newline 2. **Modify Sentence Length**  
\newline  - Either **extend or shorten** the sentence while preserving the original meaning.  
\newline  - If extending, add **clarifying details** or **restructure for better readability**.  
\newline  - If shortening, remove redundant words while retaining key scientific information.  
\newline 3. **Add or Delete Non-Essential Words**  
\newline   - Introduce or remove words **that do not alter** the fundamental meaning.  
\newline  - Improve fluency by modifying transitions or simplifying phrasing.  
\newline 4. **Alter Sentence Structure**  
\newline   - Rearrange the sentence while maintaining logical flow.  
\newline   - Convert passive to active voice (or vice versa) where appropriate.  
\newline   - Split complex sentences into simpler ones or combine shorter ones for a smoother read.  
\newline 5. **Ensure Significant Differences from the Original**  
\newline   - The paraphrased versions **must be structurally different** while still preserving the core message.  
\newline 6. **Output Format:**  
\newline   - Return **only** the elaborated subset sentences, numbered **1 to 5**, with no additional text or explanations.  
\newline \#\#\# **Example Input \& Output**  
\newline \#\#\#\# **Example 1: Scientific Context**  
\newline**Input Sentence:**  
\newline*"Long-term exposure to polluted air has been linked to an increased risk of developing respiratory illnesses such as asthma and chronic bronchitis."*  
\newline**Generated Sentences:**  
\newline1. Extended periods of contact with contaminated air are associated with a greater likelihood of respiratory conditions, including asthma and bronchitis.  
\newline2. Prolonged inhalation of polluted air may elevate the probability of experiencing chronic lung diseases.  
\newline3. Airborne pollutants have been found to contribute to the onset of various respiratory disorders over time.  
\newline4. Studies suggest that individuals consistently exposed to poor air quality are at a heightened risk of breathing-related health issues.  
\newline5. The presence of hazardous particles in the atmosphere can gradually impair lung function and lead to chronic respiratory distress.  
\newline---
\newline\#\#\#\# **Example 2: AI \& Technology Context**  
\newline **Input Sentence:**  
\newline *"Machine learning algorithms are increasingly being used to detect patterns in large datasets, improving decision-making processes in fields such as healthcare and finance."*  
\newline **Generated Sentences:**  
\newline 1. AI-driven models are enhancing decision-making by identifying trends in vast amounts of data.  
\newline 2. The application of machine learning techniques is revolutionizing data analysis across multiple industries.  
\newline 3. Advanced computational models help uncover hidden insights within extensive datasets.  
\newline 4. The use of artificial intelligence in fields like healthcare and banking is optimizing predictive analytics.  
\newline 5. Modern machine learning tools enable more efficient data-driven decisions through pattern recognition.  
\newline ---
\newline Now, generate up to {k} paraphrased sentences for the following input:  
\newline **Input:** {\color{blue}  \{idea\}}
}
\end{tcolorbox}
\end{figure*}

\begin{figure*}[t]
  \centering
  \begin{tcolorbox}[colback=yellow!10, colframe=black!89!white, before skip=5pt, after skip=5pt,
      title=\textbf{Prompt 2 (b):} Detailed prompt for partial idea generation by LLMs,
      fontupper=\small,               
      sharp corners,
      ]  
{{\color{red} System}: You are an expert in scientific writing and language transformation. Your task is to paraphrase scientific statements while ensuring clarity, precision, and natural readability. The meaning of the original sentence must remain unchanged, but the structure and expression should be significantly altered.
\newline
{\color{red} User}: 
\newline \#\#\#\#**Task:** Given an input sentence, generate up to k subset sentences that retain part of the original information while allowing for minor modifications. You may add, delete, or replace words as long as the subset sentence conveys a meaningful portion of the original content without including all of it. Each subset sentence should be **elaborated**, providing additional context or explanation while staying true to the original meaning.
\newline  \#\#\#\# **Instructions:**  
\newline 1. Extract a meaningful portion of the input sentence.  
\newline 2. Modify the extracted portion by **adding, deleting, or replacing** words while ensuring clarity and coherence.  
\newline  3. Elaborate on the extracted information by adding context, explanation, or detail while retaining the core meaning.  
\newline  4. Each subset sentence should retain **only part of the original information**, not the full meaning.  
\newline  5. Ensure grammatical correctness and natural phrasing.  
\newline  6. Avoid directly copying the exact words from the input sentence.  
\newline  7. Do not introduce entirely new or unrelated information.  
\newline  8. **Output only the elaborated subset sentences, numbered from 1 to 10, with no additional text or explanations.**
\newline --- 
\newline \#\#\# **Example Input \& Output**  
\newline \#\#\#\# **Example 1:**  
\newline  **Input:** "The increasing reliance on artificial intelligence in the healthcare industry is transforming patient diagnostics and treatment planning."  
\newline 
\newline  **Output:**  
\newline 1. AI is revolutionizing healthcare by enhancing how doctors diagnose illnesses and design personalized treatments.  
\newline 2. The healthcare sector is increasingly adopting AI-driven tools to streamline diagnostic processes and improve patient outcomes.  
\newline 3. Advanced algorithms are now assisting healthcare providers in identifying conditions more accurately and efficiently.  
\newline 4. Artificial intelligence is not only transforming diagnostics but also reshaping how treatment plans are tailored to individual patients.  
\newline 5. Medical professionals are leveraging AI systems to analyze patient data and detect abnormalities earlier.  
\newline --- 
\newline \#\#\# **Example 2:**  
\newline*"Long-term exposure to polluted air has been linked to an increased risk of developing respiratory illnesses such as asthma and chronic bronchitis."*  
\newline**Generated Sentences:**  
\newline1. Extended periods of contact with contaminated air are associated with a greater likelihood of respiratory conditions, including asthma and bronchitis.  
\newline2. Prolonged inhalation of polluted air may elevate the probability of experiencing chronic lung diseases.  
\newline3. Airborne pollutants have been found to contribute to the onset of various respiratory disorders over time.  
\newline4. Studies suggest that individuals consistently exposed to poor air quality are at a heightened risk of breathing-related health issues.  
\newline5. The presence of hazardous particles in the atmosphere can gradually impair lung function and lead to chronic respiratory distress.  
\newline---
\newline\#\#\#\# **Example 2: AI \& Technology Context**  
\newline **Input:** "Due to climate change, extreme weather events such as hurricanes and heatwaves have become more frequent and intense."  
\newline **Output:**  
\newline1. Rising global temperatures are causing hurricanes to become more powerful and destructive.  
\newline2. Climate change is intensifying heatwaves, making them last longer and reach higher temperatures.  
\newline3. Severe weather patterns are now more common due to the warming atmosphere and changing climate conditions.  
\newline4. The increased occurrence of hurricanes and heatwaves is a direct result of shifts in global weather systems.  
\newline5. Scientists link the rise in extreme weather events to the ongoing effects of climate change.  
\newline---
\newline Now, generate up to {k} elaborated subset sentences for the following input:  
\newline **Input:** {\color{blue}  \{idea\}}
}
\end{tcolorbox}
\end{figure*}

\begin{figure*}[t]
  \centering
  \begin{tcolorbox}[colback=yellow!10, colframe=black!89!white, before skip=5pt, after skip=5pt,
      title=\textbf{Prompt 2 (c):} Detailed prompt for incremental idea generation by LLMs,
      fontupper=\small,               
      sharp corners,
      ]  
{{\color{red} System}: You are an expert in scientific writing and language transformation. Your task is to paraphrase scientific statements while ensuring clarity, precision, and natural readability. The meaning of the original sentence must remain unchanged, but the structure and expression should be significantly altered.
\newline
{\color{red} User}: 
\newline \#\#\#\#**Task:** Given two input sentences, generate **up to k sentences** that retain the original information with no modifications or elaborations. The generated sentences are LAGRELY different from these two papers in semantic level (e.g., low BERT-based similarity), but bring the key ideas from them. For example, if the paper A has distinct ideas: idea\_A1, idea\_A2, and paper B distinct ideas: idea\_B1, idea\_B2, you can generate the idea by selecting subset of ideas from both these two papers (e.g., ideas idea\_A1 + idea\_B2 -> new abstract). Besides that, you can follow the below rules to help you the generated abstract that seems to be largely distinct to the original papers (e.g., Paper A and Paper B):
\newline  \#\#\#\# **Instructions:**  
\newline1. **Rephrase \& Restructure**: The new sentence must **significantly alter** the structure and wording of the selected subsets while **retaining only their core ideas**. 
\newline2. **Break Down \& Blend**: Instead of copying large portions, **extract fragments** from the subsets and recombine them in a different way. 
\newline3. **Introduce Metaphors or Analogies**: Use figurative language to convey the original meaning in a more indirect manner. 
\newline4. **Use Different Sentence Structures**: Experiment with **questions, lists, cause-effect statements, conditionals, or passive voice**. 
\newline5. **Limit Direct Keywords**: Avoid using the **exact phrasing** from the original subsets-paraphrase where possible.
\newline6. **Focus on Implicit Links**: The new sentence **should not clearly resemble either source sentence**, making it harder to trace back to any one subset. 
\newline7. **Enforce Conceptual Fusion**: The new sentence must not focus too heavily on just one of the selected subsets but should **merge ideas from both** in a way that feels natural and balanced. 
\newline8. **Ensure Consistency**: Does not create a new perspective or connection between the ideas, presents the ideas together in one sentence, but without forcing a relationship.
\newline9. **Output Format:**  
\newline  - Return **only** the elaborated subset sentences, numbered **1 to 5**, with no additional text or explanations.  
\newline --- 
\newline \#\#\# **Example Input \& Output**  
\newline \#\#\#\# **Example 1:**  
\newline**idea\_A1:** Personalization improves email engagement.
\newline**idea\_B2:** Scarcity messaging can damage long-term trust.
\newline**Generated Sentence:**
\newline1. Tailored content boosts email responses, while limited-time offers may compromise brand credibility.
\newline
\newline\#\#\#\# **Example 2: NLP Domain**
\newline **idea\_A2:** Instruction tuning aligns model outputs with user prompts.
\newline**idea\_B1:** Retrieval enhances factual accuracy in text generation.
\newline
\newline**Generated Sentence:**
\newline1. While instruction tuning improves adherence to task instructions, adding retrieval helps ground outputs in reliable information.
\newline---
\newline Now, generate up to k fused sentences for the following inputs:  
\newline Sentence A: {\color{blue} Idea 1}
\newline Sentence B: {\color{blue} Idea 2}
}
\end{tcolorbox}
\end{figure*}

\begin{figure*}[t]
  \centering
  \begin{tcolorbox}[colback=yellow!10, colframe=black!89!white, before skip=5pt, after skip=5pt,
      title=\textbf{Prompt 3:} Trigger LLMs for novelty detection.,
      fontupper=\small,               
      sharp corners,
      ]  
{{\color{red} System}: You are given a *"new research idea"* and a list of *"existing research ideas."* Your task is to compare the new idea **against each existing idea** and assign a **novelty score**.
\newline
{\color{red} User}: \#\#\# \textbf{Clarified Scoring Criteria:}
\newline Use the following **novelty scoring rubric** for each comparison:
\newline Scoring Guidelines:  
\newline 0.0 – No Novelty (Identical / Reworded). The new idea is a direct copy or a rephrased version of an existing idea.  Shares the same claims, findings, information or logic.  Example: Paraphrasing an existing idea without introducing any change.
\newline 0.3  – Low Novelty (Subset of Existing Idea). The new idea is a strict subset of an existing idea. Removes a condition, claim, or component, but otherwise shares the same logic and goal.  No new dimension or direction added.  Example: Taking just “Claim A” from “Claim A + Claim B”.
\newline 0.5  – Moderate Novelty (large partial overlap). The new idea shares a substantial portion (around 50\%) of the ideas or claims with an existing idea. It introduces some new elements, such as a different combination of claims or modified emphasis, but it’s not entirely new. It’s not a subset or superset, but has a large information intersection. Example: “Claim A + Claim C” compared to existing “Claim A + Claim B”.
\newline 0.7  – High Novelty (small partial overlap). The new idea has minor similarities (e.g., method or framing), but differs in research focus, target, or core claim. It applies known ideas in a new context or formulates a distinct question, showing clear divergence from existing ideas. Example: Using a social theory from marketing in the context of education.
\newline 1.0 – Very High Novelty (Distinct Research Direction). The idea is entirely distinct in structure, claim, scope, and research objective. Only minor thematic or terminological similarity, if any. Example: Proposing a brand new framework, theory, or dataset with no precedent in existing ideas.
\newline \textbf{\#\#\# Evaluation Instructions:}
\newline For each comparison between the new idea and an existing idea:
\newline 1. Assess Overlap. Examine the conceptual and structural overlap between the new idea and the existing idea.
\newline 2. Assign a Novelty Score. Use the Novelty Scoring Rubric to assign a score between 0.0 and 1.0, based on the degree of similarity or difference.
\newline 3. Repeat for All Comparisons. Perform this scoring for each existing idea in the set.
\newline \#\#\# \textbf{Output Format:} [List of scores like: [0.3, 0.5, 0.3, 0.7, 1.0]]
\newline Now, compare the given research idea with each of the existing ideas. For each comparison, assign a novelty score using the rubric above and no need to justify the score. Please return only a Python-style list of the novelty scores for each comparison and the final decision, no further explanation. 
\newline Given Idea: {\color{magenta}{(QUERY) The investigation discovers that the effectiveness of online ads is influenced by the volume and placement of ads, and a data-driven strategy to manage ad exposure could potentially increase campaign success by 15\%.}}
\newline {\color{blue} Existing Ideas:}
\newline {\color{blue} 1. The study finds that varying degrees of wearout and weariness exist among consumers in response to online ad volume and placement, with an appropriate "profiling and capping" strategy potentially improving ad deployment effectiveness by 15\%. }
\newline {\color{blue} ... }
\newline {\color{blue} K. The study finds that display advertisements have a low direct effect on purchase conversion but stimulate subsequent visits through other advertisement formats, and the commonly used measure of conversion rate is biased in favor of search advertisements, underestimating the conversion effect of display advertisements. }
}
\end{tcolorbox}
\end{figure*}

\end{document}